\documentclass[11pt]{article}

\usepackage{arabtex}
\usepackage[utf8]{inputenc}
\def\gap{\vspace{0.1cm}}
\newcommand{\trl}{\mbox{---}} %trattino lungo, per gli incisi

\renewcommand{\theenumi}{{\sc(}{\it\alph{enumi}\hspace*{0.5pt}}{\sc)}}

\newenvironment{numerate}%              % added on Oct. 6, 2000
{\begin{list}{\mbox{{\theenumi}}} %\hfill}%     % modified Oct. 14, 2000
{\setlength{\itemsep}{0pt}%
 \setlength{\parsep}{2pt}%
 \setlength{\leftmargin}{40pt}%          
 \setlength{\labelwidth}{25pt}%
 \usecounter{enumi}}}%
{\end{list}}

\newtheorem{example}[equation]{{\it Example: }}
\newtheorem{remark}[equation]{{\it Remark: }}

\newcommand{\br}{\begin{remark}\rm}
\newcommand{\er}{\end{remark}}
\newcommand{\be}{\begin{example}\rm}
\newcommand{\ee}{\end{example}}

\usepackage{url}

\usepackage{listings} % for JSON
\usepackage{xcolor}
\definecolor{delim}{RGB}{20,105,176}
\lstdefinelanguage{json}{
    basicstyle=\ttfamily\scriptsize,
    numbers=left,
    numberstyle=\tiny,
    stepnumber=1,
    breaklines=true,
    frame=lines,
    backgroundcolor=\color{lightgray!10},
    literate=
    {:}{{{\color{delim}:}}}{1}
    {,}{{{\color{delim},}}}{1}
    {\{}{{{\color{delim}\{}}}{1}
    {\}}{{{\color{delim}\}}}}{1}
    {[}{{{\color{delim}[}}}{1}
    {]}{{{\color{delim}]}}}{1},
}

\newcommand{\PWN}{\mbox{PWN}} %Princeton WordNet
\newcommand{\UWN}{\textsc{UzWordnet}} % Uzbek wordnet
\newcommand{\UWNshort}{\textsc{UzW}} % Uzbek wordnet (short form)
\newcommand{\KIT}{\mbox{Kitaabi}} % Kitaabi (existing)
\newcommand{\AIH}{\textsc{AI Helper}} % AI Helper
\newcommand{\GPT}{\textsc{GPT}} % GPT series
\newcommand{\CHATGPT}{ChatGPT} % ChatGPT

\usepackage[linesnumbered, ruled,noline]{algorithm2e}
\usepackage{tcolorbox}
\usepackage{setspace} % for interline spacing in tcolorboxes

\newcommand{\SFG}{\textsc{SF-game}} % Sentence-Formation Algorithm 
\newcommand{\SCG}{\textsc{SC-game}} % Spelling Corrector Game
\newcommand{\FWG}{\textsc{FW-game}} % Free Writing Game

\newcommand{\WB}{\textsc{WordBase}} % Word-Base
\newcommand{\WBshort}{\textsc{WB}} % Word-Base (short form)

\newcommand{\LANG}{\mbox{$\mathcal{L}$}} % A natural language

\newcommand{\TEXT}{\mbox{\sl TEXT}}
\renewcommand{\TEXT}{\textsf{\upshape\mdseries Text}} 

\newcommand{\PROMPT}{\mbox{\textit{Prompt}}} % generic AI ``prompt''
\renewcommand{\PROMPT}{\mbox{\textsf{\upshape\mdseries Prompt}}}
\newcommand{\CEFR}{\mbox{\textsf{\upshape\mdseries CEFR}}} 
\newcommand{\RESP}{\mbox{\bf\textsf{R}}} % response by LLM on input ``prompt''

\newcommand{\ASEN}{\mbox{$\theta$}}    % a given sentence
\newcommand{\LEM}{\mbox{$L$}}          % set of lemmas for prompt in SF-game

\newcommand{\TL}{\mbox{$t$}} % time-limit in SF-game
\newcommand{\SCORE}{\mbox{$\sigma$}} % score in SF-game
\newcommand{\KLEM}{k} % cardinality of \LEM in SF-game
\newcommand{\NSEN}{\mbox{$\delta$}} % threshold for lemmas OK in SF-game

\newcommand{\FAGENT}{\mbox{$\Psi$}}    % (formal) learner
\newcommand{\FBGENT}{\mbox{$\Phi$}}    % (formal) LLM      

\begin{document}   
\title{\UWN\ and Generative AI \\ for Learning Uzbek by Game Playing%
\thanks{%
Electronic correspondence to: a.agostini@inha.uz / agostini@disi.unitn.it.}}

\author{%
Alessandro Agostini \\ Inha University, Tashkent \and
Saydobid Khusanov \\ NUIST, Nanjing, China \and
Mirkamol Mirkamilov \\ NUIST, Nanjing, China}

\date{}
\bibliographystyle{alpha}
\maketitle
\begin{abstract}
\noindent
This paper presents an educational system architecture that enables 
learners to practice the Uzbek language through game-playing. The architecture
integrates \UWN\ and the largest currently available orthographic
dictionary for Uzbek as core lexical resources, together with generative AI  
as a fundamental component for learning support. 
We design four educational games to facilitate Uzbek language
learning and propose a game-based methodology for
improving \UWN\ as a direct by-product of game dynamics. 
Our approach combines game design and lexical resources to
address objectives that are at the same time educational (language
learning) and lexical (improvement and enrichment of a lexical resource). 
\end{abstract}

\section{Introduction}
\label{s-intro}

Game-based paradigms of learning help to penetrate cultural limits and
eventually transform them into learning advantages for the student,
since games are natural learning environments in most cultures
\cite{Huizinga49}. 
Games involve competition, death and victory, deception and
frustration, emotions \trl\ exactly as in real life. 

Learning natural languages is not an exception. This partially
explains why millions of people use ``gamified'' language learning
systems to study their preferred language
\cite{Shortt2023call,Zhang2015joep,Karasimos2022rpll}. 
It seems natural: language learning is a cognitive process, a ``mental
action'' well-expressed by playing a game. 

The question is: How game-based paradigms of language learning and
Artificial Intelligence (AI), especially the AI called ``generative''
\cite{Law2024ceo}, help the avid learner to progress?
Applications like Duolingo
\cite{Cassie2023duolingo} and others 
\cite{Alshumaimeri2024ieee-tlt,Essafi2024jct,Karasimos2022rpll,Sari2022jrt,Sofa2022taq}
provide each an answer. Unfortunately, as far as we know, none of the
existing language-learning platforms include the Uzbek language as an
option.  

In this paper, we focus on the learning of Northern Uzbek, a Turkic
language officially recognized as the national language of the Republic
of Uzbekistan. 
In particular, we build on an existing educational system architecture
for Uzbek language learning, called \KIT\footnote{``My book'' from Arabic
(\RL{kitabi}), 
symbolizing writing, knowledge, and learning.} \cite{Agostini2025STIARedit}, and
integrate its linguistic resources with \UWN\ \trl\ a lexical resource
open-source\footnote{Available at https://github.com/LDKR-Group/UzWordnet.} 
for Northern Uzbek; see for instance \cite{Agostini21ACLcoll} \trl\ and a
word base built on the largest and most complete orthographic
dictionary for Uzbek currently available \cite{Begmatov2023odictionary}.
Moreover, our development leverages \UWN, in addition to generative AI,
specifically the Generative Pre-trained Transformer 3.5 (\GPT\ 3.5 for
short; see \cite{Brown2020NeurIPS}), to implement three new features,
each in the form of a game: for spelling correction, creative writing, and
use of definitions.

The results of this paper are the following:
\vspace{-5pt}
\begin{numerate}
\item
%(1) 
We design a novel software architecture for learning Uzbek
through a game-based approach integrating \UWN\ and LLM generative AI.

\item
%(2) 
We design four educational games to facilitate Uzbek language learning.

\item
%(3) 
We provide a methodology to improve \UWN\ 
\trl\ and any lexical database, or ``word-net'', as we
generally mean the term today after the 
work started in mid 1980s that produced Princeton WordNet
\cite{Miller95cacm,Wordnet98} \trl\ as a direct by-product of game dynamics.
\end{numerate}

This paper is organized as follows. 
In the next section, 
we present a few elements of the Uzbek language. We
will see that, in spite of its official status, the language has been
experimenting with a number of issues for the disclosure of its full
potential, especially in the digital world. 
This fact has arguably made its study, learning, and teaching difficult.
In Section \ref{s-architecture},
we advance and discuss the proposed system architecture. 
In Section \ref{ss-games}, we describe the most characterizing
component of the architecture, which contains four games for Uzbek
learning. Three games also serve to enrich and improve 
the main lexical resource, namely 
\UWN, by game-playing (Section \ref{ss-enhanceUWN}).
In Section \ref{s-RW}, we discuss the related work.
In Section \ref{s-Concl+FW}, we end the paper with conclusion,
limitations, and future work.

\section{The Uzbek Language}
\label{s-Uzbek}

The Uzbek language (native: \textit{O`zbek tili}) or, more precisely,
Northern Uzbek, is the statutory language of the
Republic of Uzbekistan.
It is a Turkic language and, together with Southern Uzbek, is spoken by approximately 
34.1 
million people around the world
\cite{EthnUz2025url}, remarkably by a large group of ethnic Uzbeks
residing in Uzbekistan and abroad in Afghanistan, Kyrgyzstan,
Kazakhstan, Turkmenistan, Tajikistan, Russia, Turkey, and Xinjiang (China).
Uzbek is the second-most widely spoken Turkic language after Turkish
\cite{EthnRank2025url}.\footnote{Turkish (91.3M), Azerbaijani
  (24.2M), Kazakh (21.4M), Uyghur (13.6M).}

Unless otherwise stated, in this paper we limit our discussion to 
Northern Uzbek, and refer to it simply as Uzbek. 

\begin{figure}[!ht]
\centering

\begin{minipage}[c]{0.48\textwidth}
\vspace{0pt}
\centering
\includegraphics[width=\linewidth]{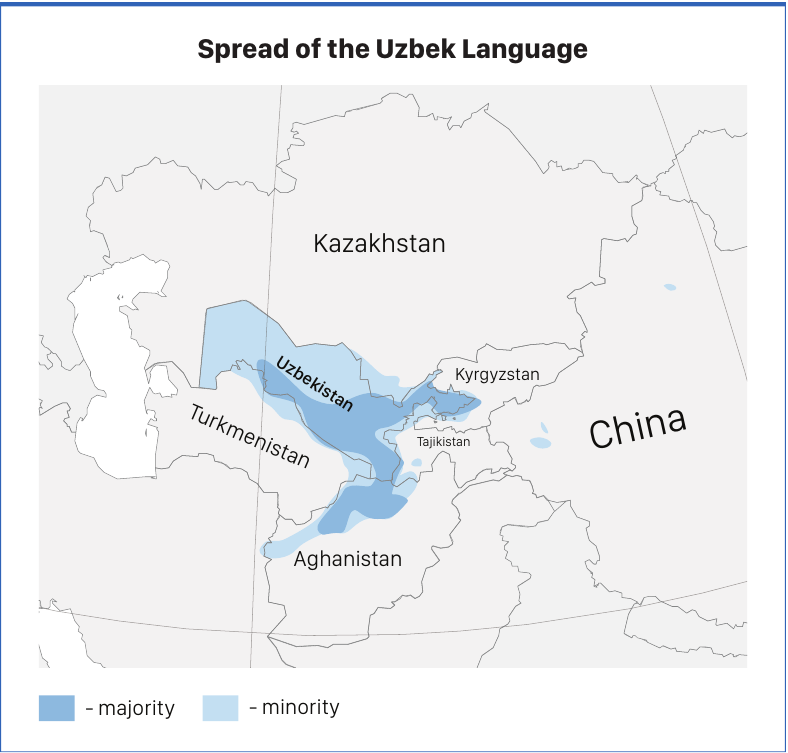}
\end{minipage}
\hfill
\begin{minipage}[c]{0.48\textwidth}
\vspace{0pt}
\centering
\includegraphics[width=\linewidth]{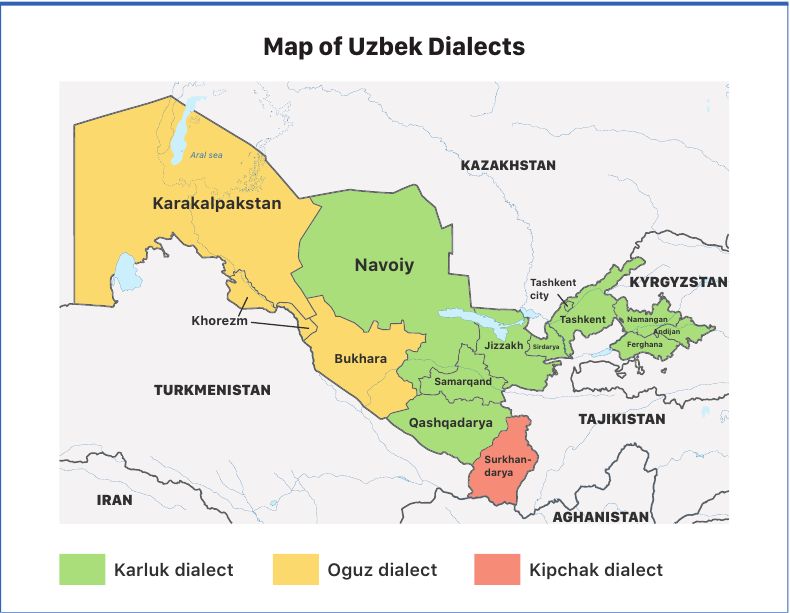}
\end{minipage}

\caption{Spread of Uzbek languages and Uzbek dialects (cf. \cite{Agostini21ACLcoll}).}
\label{fig:map}
\end{figure}

\subsection{Digital Language Support}
\label{ss-digitalsupport}

Following \cite{EthnUz2025url}, Uzbek 
is \textit{institutional} (``the language has been developed to
the point that it is used and sustained by institutions beyond the
home and community'') and \textit{vital} (``the language is supported
by multiple [digital] tools [...] it has digital contents, encoding tools,
some spell checking, some machine translation, some speech processing'';
see also \cite{Simons2022COOLING}. However, despite its vitality,
the computational development and lexical resources available for
Uzbek in the digital world are limited. 
As a result, Uzbek can be classified as a \textit{low-resource} language.

We believe that two main reasons
have delayed the development of digital support for Uzbek. 
First, computational linguistics appeared as a field of research in
Uzbekistan since the late 2000s, so relatively recently; see 
\cite{Pulatov11book,Rakhimov11book} and \cite[pp. 17-19]{Abdurakhmonova20book}.
Second, Uzbek faced a number of challenges with syntax throughout
history, with the alphabet going through changes four times in the
20th century \cite{Ryan2020cap,Shozamonov2021ojml}. 
As a consequence of the aforementioned
issues, the avid, enthusiastic, and serious learner 
of Uzbek faces numerous difficulties, such as distinguishing the
``right alphabet'' at first, or using the correct spelling, finding
reliable lexical resources, textbooks, and any 
material suitable for learning Uzbek.

\section{System Architecture}
\label{s-architecture}

\begin{figure*}[th!]  % * = twocolumn
\centering
\includegraphics[width=0.8\textwidth]{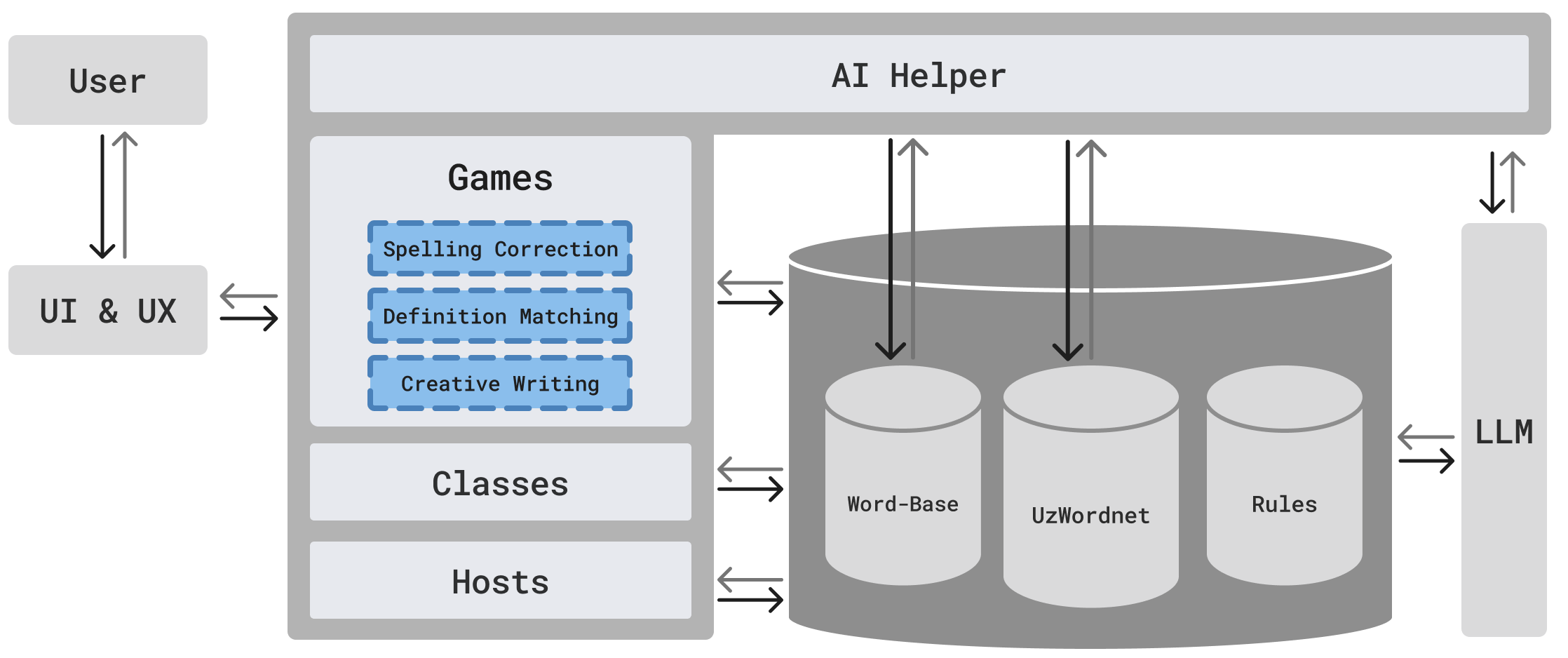}
\caption{The system architecture with \UWN.} 
\label{fig:system-architecture}
\end{figure*}

The technical focus of this paper is on system design from a software
engineering perspective. In this section, we describe the overall and specific
components of the architecture we propose, and emphasize the
connection between different functional modules. 
The components are depicted in Figure \ref{fig:system-architecture}. 

\noindent

\noindent
The main components are seven: 
AI Helper, Rules, Classes, Hosts, WordBase, \UWN, Games.% 
\footnote{\textsc{Classes} addresses the learner to
  standard lessons; \textsc{Hosts} lets teachers to organize their
  classes. These components of the system are outside the scope of this paper.} 
In the following, we describe the most important components relevant
to understanding our specific contribution. 

\gap\noindent
\textbf{\AIH.} 
This module is designed to support the system in building the structured
information (``prompt'') that instructs the language model in use to
generate what is necessary for the specific educational or lexical
objective. 

Since the objectives pursued in this paper are achieved
through game-playing, the \AIH\ is defined differently depending on the
specific game design. Accordingly, we postpone its detailed description
to Section \ref{ss-games}.

\gap\noindent
\textbf{\textsc{Rules}.}
To provide grammar content at the elementary and intermediate levels,
we adopt the CEFR Rules \cite{CoE2020cefr}, which are stored in this
component. Grammar rules are used by the \AIH\ to operate in one game
(Game 0, Section \ref{ss-games}). 

\gap\noindent
\textbf{\WB.}
The first lexical module we present 
stores linguistic data used in the game's modules 
\textsc{Spelling Corrector} and \textsc{Sentence Formation} (see
Section \ref{ss-games}). 
We built the component by using an orthographic dictionary of Uzbek
words, specifically \cite{Begmatov2023odictionary}, as well as two
sets of English words, exactly \textit{The Oxford 3000 by CEFR 
  level} \cite{Oxford2019cefr}%
\footnote{The list contains the most common 3000 English words categorized by the CEFR levels.}, 
 and the \textit{A2 Key and B1 preliminary vocabulary
  lists} by Cambridge University Press and Assessment \cite{Cambridge2019A2key}.%
\footnote{The lists contain the words and topics that mostly occur in Cambridge Assessment tests.} 

\br
\label{rmk-vocabulary}
Since the proposed architecture \trl\ specifically, the design of Game 0
(first game in Section \ref{ss-games}) \trl\ is intended to support the
learning of Uzbek at beginner, elementary, or intermediate levels 
(i.e., CEFR levels A1, A2, and B1, respectively; \cite[sec. 3.3]{CoE2020cefr}), 
the WordBase is expected to store Uzbek words classified 
according to these levels.
However, to the best of our findings, there are no resources available
mapping the Uzbek language into A1, A2, B1 levels. Therefore, we decided to
store English words and CEFR levels as a reference to the most
``useful'' or ``common'' Uzbek words, \emph{after
translation from English} of those words. The working assumption we
have done here presumes that ``the most common words'' of a language are 
the same across all languages and cultures.
\er

\gap\noindent
\textbf{\UWN.}
\label{ss-UWN}
We employ a lexical database for Uzbek to provide the necessary
lexicon in support of game design.
We opted for \UWN\ \cite{Agostini21ACLcoll} because it is
open-source%
\footnote{https://globalwordnet.github.io/resources/wordnets-in-the-world. Accessed: 2026-02-21.}
and currently the largest wordnet for Uzbek, and the second
wordnet developed overall.\footnote{The first is part of the Extended
  Open Multilingual WordNet project \cite{Bond13ACL}.} 
\UWN\ is compatible with semantic networks, or ``word-nets'', of the kind
first developed with 
Princeton WordNet (\PWN\ for short) \cite{Miller95cacm,Wordnet98}. Its
use allows us to possibly extend our game-based methodologies to other
languages linked through the Open Multilingual Wordnet
\cite{Bond12GWC,Bond13ACL}. 

\UWN\ was developed by expanding the semantic structure 
of \PWN\ according to the \textit{extend method} \cite{Bond12GWC}, 
which assumes that the lemmas of 
the target language are created 
from English synsets of \PWN.

\section{Games} 
\label{ss-games}

The last and most characterizing component of the architecture
is 
\textsc{Games}; it contains four game-based modules. We now
describe each of them in turn.  

\gap\noindent
\textbf{\textsc{Game 0 -- Free Writing}.} 
By playing this game, a learner can verify the quality of her writing
and receive feedback. 
We call this game the \textit{Free Writing Game} (\FWG\ for short).
We now describe its mechanics.

The learner (Player 1) starts the game by writing some text in Uzbek;
let \TEXT\ denote it.
The system (Player 2) sends a request to the \AIH\ to build a ``prompt''
(hereafter, \PROMPT; see the box below for details) upon \TEXT.
Successively, the system sends \PROMPT\ to a suitable large language model (LLM
component in Figure \ref{fig:system-architecture}). In this paper, we
refer to \GPT\ 3.5
via its standard API.
The \GPT\ series has provably shown outstanding performance in
unifying all natural language understanding tasks into generative
tasks \cite{Brown2020NeurIPS}, so we assume it is well-suitable
as a starting point for language learning. 

\noindent
Finally, the system processes the response \RESP\ (in JSON format)
from the language model and displays it to the learner 
as feedback. % for learning. 

\begin{tcolorbox}[colback=gray!5,colframe=gray!50,coltitle=black,title=\PROMPT\
    for \FWG,label={box-AIHprompt}] 
\begin{lstlisting}[language=json,numbers=none,escapeinside={*}{*}]]
You are provided with a text in language *\LANG *:
1. Set language *\LANG * to "language" Key. 
2. Set text to "originalText" Key. 
3. Find the mistakes in text and correct them. 
   Set corrected text to "correctedText" Key.  
4. Identify CEFR level of text by using rules *\CEFR *.
   Set CEFR level to "cefrLevel" Key. 
5. Provide suggestions in an array to improve writing proficiency in text.
   Set your suggestions to "suggestions" Key.   
6. Provide points on a scale of 100 for text in following criteria:
   1. Correctness: set "correctness" Key
   2. Clarity: set "clarity" Key
   3. Delivery: set "delivery" Key
7. Your response must be in JSON format!
Input: *\TEXT\ *
\end{lstlisting}
\end{tcolorbox}\noindent

Here are some examples.

\be
\label{ex-AIH1}
{\normalfont [\TEXT\ \textbf{is correct}]} Suppose that \TEXT\ is:
``\textit{Yurtimizga xush kelibsiz. Sizni ko`rganimizdan mamnunmiz}''
(``Welcome to our country. We are pleased to see you.'').
Then:

1. \PROMPT\ is built by \AIH\ and sent to LLM component of the system. 

2. The language model generates the response \RESP\ (in JSON format),
namely:

\gap 
\begin{lstlisting}[language=json,numbers=none]
{
  "language": "Uzbek",
  "originalText": "Yurtimizga xush kelibsiz. Sizni ko`rganimizdan mamnunmiz.",
  "correctedText": "Yurtimizga xush kelibsiz. Sizni ko`rganimizdan mamnunmiz.",
  "cefrLevel": "A2",
  "suggestions": [
    "Ravonlik va ta'sirchanlikni oshirish uchun murakkabroq gap tuzilmalaridan foydalaning.",
    "Sinonimlardan foydalanib, so`z boyligingizni oshirishga harakat qiling.",
    "Rasmiy yoki norasmiy kontekstga mos ravishda muvofiq so`z tanlashni o`rganing.",
    "Tinish belgilari va ifodaviy vositalardan to`g`ri foydalanish bilan matnning ta'sirchanligini oshiring."
  ],
  "correctness": 100,
  "clarity": 100,
  "delivery": 95
}
\end{lstlisting}

3. The system decodes \RESP\ and displays via user interface the feedback:
\\[3pt] 
$<Correctness = 100\%, Clarity = 100\%, Delivery = 95\%>$
\\[3pt] 
together with CEFR's level (A2) and suggestions to improve \TEXT. 
\ee\

\br\
By CEFR's rules, if \TEXT\ is too simple or short, then
the only levels available are A1 (``Beginner''), and A2
(``Elementary''); see \cite[sec. 3.3]{CoE2020cefr} for the definition
of A1 and A2 language proficiency levels. 
\er\

\be
\label{ex-AIH2}
{\normalfont [\TEXT\ \textbf{is incorrect}]} Suppose \TEXT\ is:
``\textit{Sizzi ko`rganimdan hursanman. Axvollariz qaley?}''
(``I am happy to see you. How are you?'') Then:

1. \PROMPT\ 
is built by \AIH\ and sent to LLM component of the system.

2. The language model generates the response \RESP\ (in JSON format), namely:

\gap 
\begin{lstlisting}[language=json,numbers=none]
{
  "language": "Uzbek",
  "originalText": "Sizzi ko`rganimdan hursanman. Axvollariz qaley?",
  "correctedText": "Sizni ko`rganimdan xursandman. Ahvolingiz qanday?",
  "cefrLevel": "A2",
  "suggestions": [
    "Imlo va grammatik jihatdan to`g`ri yozishga e'tibor bering.",
    "Rasmiy va norasmiy uslub o`rtasidagi farqni yaxshiroq tushunish uchun misollar ustida ishlang.",
    "Talaffuz va yozuvdagi aniqlikka erishish uchun so`zlarning to`g`ri shakllaridan foydalaning.",
    "Muloqotda yanada tabiiyroq va ravon ifodalanish uchun gap tuzilmalarini kengaytiring."
  ],
  "correctness": 85,
  "clarity": 90,
  "delivery": 88
}
\end{lstlisting}

3. The system decodes \RESP\ and displays via user interface the feedback:
\\[3pt] 
$<Correctness = 85\%, Clarity = 90\%, Delivery = 88\%>$
\\[3pt] 
together with CEFR's level (A2) and suggestions to improve \TEXT. 
The correct(ed) sentence: ``\textit{Sizni ko`rganimdan  xursandman. Ahvolingiz
qanday?}'' is also displayed. 
\ee\

\be
\label{ex-AIH3}
{\normalfont [\TEXT\ \textbf{is longer and more complex}]} Suppose \TEXT\ is:
``\textit{O`zbekiston Markaziy Osiyodagi davlatdir. U yerda juda ko`p
  e'tiborga sazovor maskanlar mavjud. Bu yurtda ilm-fan rivojiga hissa
  qo‘shuvchi ko`plab olimlar istiqomat qilishadi.}''
(``Uzbekistan is a country in Central Asia. There are many well-known
places there. Many scientists who contribute to scientific
advancements dwell in this country.'')
Then:

1. \PROMPT\ 
is built by \AIH\ and sent to LLM component of the system.

2. The language model generates the response \RESP\ (in JSON format), namely:

\gap 
\begin{lstlisting}[language=json,numbers=none]
{
  "language": "Uzbek",
  "originalText": "O`zbekiston Markaziy Osiyodagi davlatdir. U yerda juda ko`p e'tiborga sazovor maskanlar mavjud. Bu yurtda ilm-fan rivojiga hissa qo`shuvchi ko`plab olimlar istiqomat qilishadi.", 
  "correctedText": "O`zbekiston Markaziy Osiyoda joylashgan davlatdir. U yerda e'tiborga sazovor joylar juda ko`p. Bu yurtda ilm-fan rivojiga hissa qo`shayotgan ko`plab olimlar yashaydi.", 
  "cefrLevel": "B1",
  "suggestions": [
    "Ravonroq ifoda uchun 'Markaziy Osiyodagi' o`rniga 'Markaziy Osiyoda joylashgan' iborasidan foydalanish yaxshiroq bo`lishi mumkin.",
    "'Maskanlar' o`rniga 'joylar' so`zi ishlatilsa, gap yanada tabiiy eshitiladi.",
    "'Istiqomat qilishadi' o`rniga 'yashaydi' so`zi rasmiy va aniqroq ifoda beradi.",
    "Matnni yanada boyitish uchun murakkab gap tuzilmalaridan foydalanishni sinab ko`ring."
  ],
  "correctness": 95,
  "clarity": 92,
  "delivery": 90
}
\end{lstlisting}

3. The system decodes \RESP\ and displays via user interface the feedback:
\\[3pt] 
$<Correctness = 95\%, Clarity = 92\%, Delivery = 90\%>$
\\[3pt] 
together with CEFR's level (B1) and suggestions to improve \TEXT.
\ee\

\br
By CEFR's rules, if \TEXT\ is sufficiently long and complex, 
the possible proficiency levels are B1 (``Intermediate'') or higher (B2, C1,
C2); see \cite[sec. 3.3]{CoE2020cefr}.  
\er

\br
It is interesting to report the suggestions provided by response
\RESP\ in Example \ref{ex-AIH3}. 
Below we list such suggestions (our translation to English from Uzbek). 

\gap
\begin{lstlisting}[language=json,numbers=none,escapeinside={*}{*}]
- For smoother expression, it might be better to use *\textit{Markaziy Osiyoda joylashgan}* instead of *\textit{Markaziy Osiyodagi}*. 

- If the word *\textit{maskanlar}* is replaced with *\textit{joylar}*, the sentence sounds more natural.

- By using *\textit{yashaydi}* instead of *\textit{istiqomat qilishadi}* provides a more formal and precise expression. 

- To enrich the text, try to use more complex sentence structures.
\end{lstlisting}
\er\

\gap\noindent
\textbf{\textsc{Game 1 -- Spelling Corrector}.}
This gamified functional module designs a user session as a 2-player
game with a fixed time limit. 
We call this game the \textit{Spelling Corrector Game} (\SCG\ for
short). Here are the mechanics.

The system (Player 1) starts the game by randomly selecting a synset
$S$ from \UWN\ and a lemma $l$ in $S$. 
It then possibly modifies $l$ by one, two, or more letters and asks the 
learner to (a) guess whether $l$ is correct and, if $l$ is not
correct, (b) provide the correct lemma.  
The learner (Player 2) does answer a lemma, say $l'$. 
The system spell-check $l'$ by using 
the orthographic dictionary stored in the \WB, 
and assign a score \SCORE\ to the learner for the quality of the answer. 
To perform these two steps, the system sends a request to the \AIH\ in
order to build the 
prompt that instructs the language model to spell-check $l'$
and generate the score. Below is the prompt produced.

\begin{tcolorbox}[colback=gray!5,colframe=gray!50,coltitle=black,title=\PROMPT\ 
    for the \SCG,label={box-CRprompt},before upper={\setstretch{1.25}},after upper={}]
\begin{lstlisting}[language=json,numbers=none,escapeinside={*}{*}]]
Generate a JSON response on following criteria:
1. Lemmas: Provide an array with lemma *$l'$*.
2. Find the mistakes in *$l'$* and correct them using the *\WB*. 
3. Score: Provide a numerical score *\SCORE * (on a scale of 1 to 10)
indicating how much *$l'$* differs from correct lemma in *\WB*.
3.1 A score of 10 means that *$l'$* is correct. Lower scores are computed
proportionally to the number of letters wrong in *$l'$*.
4. Set corrected lemma to "correctLemma" Key. 
5. Set score to "correctness" Key. 
Input: *$l'$ *
\end{lstlisting}
\end{tcolorbox}\noindent
Finally, the system displays the correct lemma and the assigned score
\SCORE\ to the learner as a summary of the play session.
The learner can repeatedly play a new session starting from another
randomly selected synset and lemma, and so on. 

\br
\label{rmk-enrichingGame1}
The correct lemma is added to $S$ in place of $l$,
thereby improving \UWN\ by correcting orthographic errors in its
lemmas (if any occur). 
\er

\be
\label{ex-SpellCorr}
Let \textit{yaxshi} (``good'') be a lemma randomly retrieved from \UWN.
Suppose the system plays \textit{yahshi} by modifying
\textit{yaxshi} in one letter: \textit{x} $\rightarrow$ \textit{h}. The new lemma, 
\textit{yahshi}, is incorrect. The learner is asked to determine if
\textit{yahshi} is correct and to provide the correct lemma:
\textit{yaxshi}. If the learner succeeds, a score in the upper
range of [1,10] is assigned to the learner for the play session; otherwise,
a score in the lower range of [1,10] is assigned (see \PROMPT\ in box
above, point 3). If the game is repeated, the final score 
is computed from the  
scores assigned in all $n$ plays ($n$ sessions, a game parameter).
\ee\

\gap\noindent
\textbf{\textsc{Game 2 -- Matching Definitions}.} 
The structure of this gamified functional module is similar to that
of the previous game; it models a user session as a 2-player
game with a fixed time limit. 
However, two distinct `modes of play' are designed.
Each mode represents a different level of difficulty
for the task proposed to the learner: `easy' and  `hard'. 
We briefly explain each mode. 

Let $D$ denote a definition (``gloss'') in a synset $S$ of
\UWN.\footnote{A synset is a set of lemmas with the same
  meaning in a specific sense, cf. \cite{Miller95cacm,Wordnet98}.} 

\begin{itemize}
\item
\textbf{Easy Mode}: 
The system (Player 1) starts the game by displaying $D$ and a set of
related lemmas $l_1,... l_k$ in synset $S$ randomly selected from
\UWN\ ($k$ is a game parameter), plus a fixed number of randomly selected
lemmas in \UWN\ but not in $S$. To keep things simple, in this paper
we assume $k=1$ and define $l = l_1$.
The system then asks the learner to select $k$ lemmas
(one lemma, by assumption on $k$) that best fit the definition $D$. 
We know that such a lemma is $l$. 
The learner (Player 2) selects a lemma, say $w$. 
The system compares $w$ against each lemma in $S$.
Two cases arise:

\gap
\textit{Case 1}. A string match of $w$ with an existing lemma in $S$ is computed,
i.e., $w = l$ upon the assumption on $k$. Then, a positive score is
assigned to the learner. Moreover, one or more ``example sentences'' for $w$ are
possibly provided to the learner as feedback for 
learning. These sentences are retrieved from the synset $S$, if
such sentences are available.\footnote{Typically, a synset in a
  wordnet \emph{might} contain sentences aimed to illustrate
the usage of a lemma in the synset; these sentences are called
\textit{example sentences} in the wordnet-related literature. However,
due to incompleteness,
not all synsets in a wordnet provide example sentences. Moreover,
not all lemmas in a synset are generally associated with an example
  sentence. This may be the case for $w$.} 
If it is not available, the system sends a request to the \AIH\ in
order to build the structured information 
(``prompt'') that instructs the language model (\GPT\ 3.5 in this
paper) to generate one or more sentences using $w$, and displays the
sentences to the learner as feedback. 

\br
\label{rmk-enrichingGame2}
Contextually, each generated sentence is added
to $S$, thereby enriching \UWN\ with example sentences for $l$.
\er

\textit{Case 2}. A string match of $w$ with a lemma in $S$ does not exist.
This means that $w$ is not semantically equivalent to $l$ (or to any
$l_1,... l_k$ in $S$ in the general case).
A negative score is assigned to the learner. 
% NEW 23 FEB 2026:
Moreover, the system sends a request to the \AIH\ to build
the structured information (``prompt'') that instructs the language
model to input $<\mbox{$D$, $w$}>$ and generate feedback on the
semantic relationship, if any, between $D$ and $w$; the system then displays
the feedback to the learner.

\vspace{0.25cm}
\item
\textbf{Hard Mode}: 
The system starts the game and displays $D$. No additional information or 
guidelines are provided to the learner.
The learner writes a word (lemma) $w$ that satisfies $D$. 
The system compares $w$ against each lemma in $S$. Two cases arise,
exactly as in the Easy Mode.
\end{itemize}

\be
\label{ex-EasyMode} 
Assume that the game is played in \textbf{Easy Mode}.
Suppose the system selects the following synset $S$ from \UWN,
represented in JSON format.
\begin{lstlisting}[language=json,numbers=none]
{
  "@id": "uzwordnet-7768694-n",
  "partOfSpeech": "noun",
  "relations": [
    {
      "relType": "hyponym",
      "target": "uzwordnet-7705931-n"
    }
  ],
  "members": [ "w8801" ],
  "definition": [
    {
      "gloss": "qattiq jigarrang-qizil qobig`ida suvli qizil pulpa
      bo`lgan ko`plab urug`larga ega bo`lgan yirik sharsimon meva" 
    } ]
}
\end{lstlisting}
Note that the definition (``gloss'') in the box above means: ``A large
spherical fruit with a hard reddish-brown rind, juicy red pulp, and many seeds.'' 
Let $D$ denote such a definition.

Now suppose the system selects a lemma related to the definition,
namely, the only lemma existing in 
the synset $S$, identified by ID w8801. The lemma is \textit{anor}
(pomegranate).\footnote{The connection between the synset $S$ and the
  location of \textit{anor} is structural in \UWN. For the sake of brevity, in
  this paper we omit the technical details of this connection; see
  \cite{Agostini21ACLcoll,Miller95cacm,Wordnet98}.} 
Moreover, the system selects two more lemmas in \UWN\ but \textit{not}
in $S$. In this example, we consider these lemmas:

\gap
w4536  \textit{tuproq} (soil)

w3575 \textit{poyga} (race)

\gap\noindent
(w4536 and w3575 are IDs from \UWN, cf. \cite{Agostini21ACLcoll}). 
At this point, the system displays to the learner the tuple:
\[
<\mbox{$D$, \textit{anor}, \textit{tuproq}}, \textit{poyga}>
\]
(modulo random re-ordering)
and asks the learner to choose the lemma that best fits $D$. We know
the lemma is \textit{anor}. Let $w$ denote the learner's answer. 

If $w$ is equal to \textit{anor}, then a positive score is assigned to the
learner. 
However, an ``example sentence'' using \textit{anor} cannot be provided
to the learner by retrieving the sentence directly from $S$
(cf. JSON code above, where there are no example sentences). Therefore, the
system requires the \AIH\ to form the structured information,
or ``prompt'', that the language model inputs to generate one or more
sentences using $w$ (\textit{anor}). These sentences are finally
displayed by the system  to the learner as feedback. In addition, each
generated sentence is added to $S$, so enriching \UWN\ with example
sentences for \textit{anor}. 

If $w$ is not equal to \textit{anor}, then a negative score is
assigned to the learner. Moreover, the system performs a
kind of ``double-check'' on the pair $<\mbox{$D$, $w$}>$ to eventually
provide further feedback to the learner. 
\ee\

\be
\label{ex-HardMode}
Assume that the game is played in \textbf{Hard Mode}.
Suppose the system selects the following synset $S$ from \UWN, represented in JSON format.
\begin{lstlisting}[language=json,numbers=none]
{
  "@id": "uzwordnet-9906538-n",
  "partOfSpeech": "noun",
  "relations": [
    {
      "relType": "hyponym",
      "target": "uzwordnet-9614684-n"
    }
  ],
  "members": [ "w3394", "w4613", "w4614", "w9771" ],
  "definition": [
    {
      "gloss": "biron bir sabab uchun kurashadigan kishi" 
    }
  ]
}
\end{lstlisting}
Note that the definition (``gloss'') in the box means:``A person who fights for a cause.'' 
Let $D$ denote such a definition. Observe that the lemmas in $S$ are identified by
w3394, w4613, w4614, and w9771; in \UWN, these IDs refer to these Uzbek terms
(the corresponding lemma in English is given in parentheses):

\gap
w3394 \textit{askar} (soldier)

w4613 \textit{jangchi} (warrior)

w4614 \textit{kurashuvchi} (fighter)

w9711 \textit{paladin} (paladin, i.e. a very brave knight).

\gap\noindent
The system asks the learner to write a word that best fits $D$.
Let $w$ denote the learner's answer. 
If $w$ is equal to any of the lemmas in $S$, say $l$, then a positive score is
assigned to the learner. A negative score is assigned otherwise. The
mechanism aimed to provide the learner with ``example sentences'' for
$l$, as well as the way \UWN\ is enriched with example sentences AI-generated
to explain the use of $l$, are identical to those in Easy Mode.
\ee\

\gap\noindent
\textbf{\textsc{Game 3 -- Sentence Formation}.} 
This functional module designs a 2-player game with a fixed time
limit between the system
(Player 1; \FAGENT) and a learner (Player 2; \FBGENT) who aims to
learn how to correctly write sentences in Uzbek. We call this game the
\textit{Sentence Formation Game} (\SFG\ for short) and it is %by far
the most complex of the games proposed in this paper. 

The first design objective is educational: it provides an 
environment in which the learner can practice and improve writing 
skills. This educational feature of the game supports the ``creative
writing'' module of the architecture (Figure \ref{fig:system-architecture}). 
The second design objective, shared with Game 1 (cf. Remark
\ref{rmk-enrichingGame1}) and Game 2
(cf. Remark \ref{rmk-enrichingGame2}), is lexical: it enables the
improvement and enrichment of \UWN\ with corrected lemmas and new
example sentences.  
In the following, % paragraphs, 
we explain the game in detail (cf. also Algorithm \ref{alg-SFG}).

\begin{algorithm}[h]
\SetKwData{UzWordNet}{\UWN\ (\UWNshort\ in short)}
\SetKwData{UW}{\UWN}
\SetKwData{W}{W} 
\SetKwData{WordBase}{\WB\ (\WBshort\ in short)}
\SetKwData{WBS}{\WB}

\SetKwData{TimeLimit}{\TL}
\SetKwData{Lemmas}{\LEM}
\SetKwData{Klemmas}{\KLEM}
\SetKwData{Nsentences}{\NSEN}
\SetKwData{Sentence}{\ASEN}
\SetKwData{Score}{\SCORE}
\SetKwData{Learner}{\FAGENT}
\SetKwData{LLM}{\FBGENT}
\SetKwData{CorrectedSentence}{$\Sentence'$}
\SetKwData{UsedLemmas}{used\_lemmas}
\SetKwData{Feedback}{feedback}

\SetKwFunction{SelectLemmas}{select\_lemmas}
\SetKwFunction{StartTimer}{start}
\SetKwFunction{CheckSpelling}{spell\_checking}
\SetKwFunction{Flemmas}{lemmas}
\SetKwFunction{EvaluateSentence}{eval}
\SetKwFunction{UpdateUzWordNet}{update\_Uzwordnet}

\SetKwInOut{Input}{Input}
\SetKwInOut{Output}{Output}

\SetKwFor{For}{for}{do}{end}
\SetKwIF{If}{ElseIf}{Else}{if}{then}{else if}{else}{end}
\SetKwFor{While}{while}{do}{}

\Input{\UzWordNet, \\ \WordBase, \\ \TimeLimit\ (time-limit),
  \\ \KLEM\ (number of lemmas to select) %\;} 
  \\ $\NSEN \in [1,10]$ (threshold for suitable lemmas)\;} 

\Output{$\Score \in [1,10]$, \W\ (\UW\ updated)\;}

\SetKwProg{Fn}{Function}{:}{}

    \Lemmas $\leftarrow$ \SelectLemmas(\UW, \Klemmas, \Nsentences)\;
    
    \StartTimer(\TimeLimit)\;

    Display \Lemmas to learner \Learner\;
    
    Prompt \Learner\ to form a sentence \Sentence\ using some or all
    lemmas in \Lemmas\;

    Inform \Learner\ on higher score for more lemmas in \Lemmas\ used in \Sentence\;

    \Sentence $\leftarrow$ \FAGENT(\LEM)\;

    \CorrectedSentence$\leftarrow$ \CheckSpelling(\Sentence,\WBS)\;

    \UsedLemmas$\leftarrow \emptyset$\;
    
   \For{\bf each $l \in \Lemmas$}{
	\If {\mbox{$l \in \Flemmas(\CorrectedSentence)$}}{
	  \UsedLemmas $\leftarrow$ \UsedLemmas $\cup\, \{ l \}$}			
	}

    \Score $\leftarrow$ \EvaluateSentence(\CorrectedSentence,\UsedLemmas)\;

    \Feedback $\leftarrow$
    \LLM(\CorrectedSentence,\UsedLemmas,\PROMPT\ 2)\;

    \W $\leftarrow$ \UW\;

    \W $\leftarrow$ \UpdateUzWordNet(\W,\UsedLemmas, \CorrectedSentence)\;

\KwRet{\Score, \W}\;

\caption{\SFG\ for learner \FAGENT\ based on LLM \FBGENT.}
\label{alg-SFG}
\end{algorithm}\noindent

\gap\noindent
1. \textbf{[System Move]} \FAGENT\ randomly
selects a set $\{S_1,... S_{\KLEM}\}$ ($\KLEM$ game parameter)
of synsets from \UWN\ and one lemma $l_i$ from each $S_i$. Then \FAGENT\ 
displays the set of lemmas $\LEM = \{l_1,... l_{\KLEM}\}$ to \FBGENT\ (the learner), and
asks \FBGENT\ to: (a) guess whether each $l_i$ in \LEM\ is correct and,
if not, provide the correct lemma; and (b) write a
meaningful Uzbek sentence using one or more lemmas in \LEM,
corrected if necessary. 
Moreover, \FAGENT\ informs \FBGENT\ that the more lemmas in \LEM\ are
correctly corrected, the higher the score assigned, %for the play, 
provided that the sentence is meaningful and grammatically correct.   

\gap\noindent
2. \textbf{[Learner Move]} \FBGENT\ writes a sentence \ASEN\ using
$\LEM' = \{l'_1,... l'_{j}\}$ lemmas ($j \le \KLEM$).

\gap\noindent
3. \textbf{[System Move]} \FAGENT\ spell-checks \ASEN\ by using 
the orthographic dictionary stored
in the \WB, 
and assigns a score \SCORE\ to \FBGENT\ for the quality of
\ASEN. 

\begin{algorithm}[h]
\SetKwData{WordBase}{\WB} 
\SetKwData{WBS}{\WB}
\SetKwData{Sentence}{\ASEN}
\SetKwData{LLM}{\FAGENT}
\SetKwData{CorrectedSentence}{$\Sentence'$}

\SetKwFunction{Flemmas}{lemmas}

\SetKwFor{For}{for}{do}{end}
\SetKwIF{If}{ElseIf}{Else}{if}{then}{else if}{else}{end}
\SetKwFor{While}{while}{do}{}

\SetKwProg{Fn}{Function}{}{}

\Fn{\CheckSpelling{\Sentence, \WBS}}{

    \CorrectedSentence $\leftarrow$ \LLM(\Sentence)\;
   
    \For{\bf each $l \in \Flemmas(\CorrectedSentence)$}{
       
      \If {\mbox{$l \notin \WBS$}}{
	  \CorrectedSentence $\leftarrow$ \LLM(\Sentence,\WBS)}
    }
}
\KwRet{\CorrectedSentence}\;

\caption{Spell-checking of sentence \ASEN\ by system \FAGENT.}
\label{alg-SFG-spellchecking}
\end{algorithm}\noindent
To perform these steps, \FAGENT\ sends a request to the \AIH\ in
order to build the structured information 
(``prompt'') that instructs the language model 
to spell-check each lemma in \ASEN\ used by \FBGENT\ 
and generate the score. The prompt is described below.

\begin{tcolorbox}[colback=gray!5,colframe=gray!50,coltitle=black,title=\PROMPT\ 
    for the \SFG,label={box-CRprompt},before upper={\setstretch{1.25}},after upper={}]
\begin{lstlisting}[language=json,numbers=none,escapeinside={*}{*}]]
Generate a JSON response on following criteria:
1. Lemmas: Provide an array of lemmas *$\LEM'$ *.
2. Relation: Describe how lemmas in *$\LEM'$ * are conceptually related
in *\ASEN * and how they interact to form *\ASEN *'s meaning.
3. Find and correct errors in each lemma in *$\LEM$* using the *\WB*. 
   Denote by *$\LEM^c = \{l^c_1,... l^c_{\KLEM}\}$* the set of lemmas in *$\LEM$* corrected in 3.
4. Find and correct errors in each lemma in *$\LEM'$* using the *\WB*. 
5. Sentence Score: Provide a score *$\SCORE \in [1,10]$* measuring the
orthographic correctness of lemmas in *$\LEM'$* and their usage in *\ASEN *. 
5.1 A score of 10 means that *$\LEM' = \LEM^c$*, all lemmas in
*$\LEM'$ * appear in *\ASEN *, and the lemmas are highly related and easily combined. 
5.2 Lower scores indicate weaker usage or orthographic errors in the lemmas used in *\ASEN*. 
6. Correct lemmas: Provide an array of corrected lemmas *$\LEM^c$*.
7. Set score to "correctness" Key. 
Input: *\LEM, $\LEM'$, \ASEN *
\end{lstlisting}
\end{tcolorbox}\noindent
Finally, the system displays
the correct(ed) lemmas, the assigned score \SCORE, and an explanation of
how the lemmas used in \ASEN\ are 
related to each other and how they interact to form \ASEN's meaning
(see \PROMPT\ in box above, points 6,5,2).  

\br
\label{rmk-enrichingGame3}
Each lemma $l^c_i \in \LEM^c$ is added to $S_i$ in place of
$l_i$, thereby improving \UWN\ by correcting orthographic errors in
its lemmas (if any occur). 
\er

\section{Enhancing \UWN} 
\label{ss-enhanceUWN}

At this point, it is important to emphasize that the purpose of three
of the four educational games we defined and discussed is twofold:
educational \emph{and} lexical. 
In particular, Game 1 improves \UWN\ by correcting orthographic errors
in its lemmas (if any occur), while Game 2 (\textit{Case 1}) and Game 3 enable the
enrichment of the lexical resource employed with example sentences as
a direct consequence of the learner playing the game. 

In Game 1, the simplest of our games, lexical enrichment is performed
without constraints. Each time a game session is played, one
lemma in \UWN\ (randomly selected by the system) is either verified as
orthographically correct or replaced with a corrected version.

In Game 2, lexical enrichment is performed if and only if (a) the
learner selects lemmas in a synset $S$ of \UWN\ from among the lemmas
displayed by the system, and (b) $S$ does not contain any example
sentences for any of the lemmas the learner selected. In this case, we
have seen, the system sends a request to the \AIH\ in order to build
the prompt that instructs the language model to generate one or more
example sentences for lemmas $l_1,l_2,...$ in $S$ selected by the
learner. Here is the prompt: 

\begin{tcolorbox}[colback=gray!5,colframe=gray!50,coltitle=black,title=\PROMPT\ for the \SFG\ ($n$ game parameter),label={box-CRprompt},before upper={\setstretch{1.25}},after upper={}]
\begin{lstlisting}[language=json,numbers=none,escapeinside={*}{*}]]
Generate a JSON response on following criteria:
1. Lemmas: Provide an array of the given lemmas *\LEM *.
2. Sentences: Construct *$n$* sentences in Uzbek, each incorporating as many of the given lemmas as possible.
3. Relation: Describe how the lemmas are conceptually related and how they interact in meaning.
Input: *\LEM *
\end{lstlisting}
\end{tcolorbox}

\noindent
The prompt is used to display the $n$ example sentences
(\PROMPT, point 2) to the learner as educational feedback.
More importantly, for the focus of this section, the
sentences are added to the synset $S$ to enrich its set 
of example sentences (cf. Algorithm \ref{alg-SFG-updateUW}).
This fulfills the lexical goal of game-playing.

\gap
In Game 3, lexical enrichment is performed if and only if the score
\SCORE\ assigned to a learner for a Uzbek sentence
\ASEN\ formed using lemmas
$\LEM = \{l_1,... l_{\KLEM}\}$ ($\KLEM$ game parameter) is `good enough'
according to a game threshold.
In this case, the system (a) spell-checks the lemmas in \LEM\ using
the orthographic dictionary stored in the \WB, and (b) adds
\ASEN, corrected with the spell-checked lemmas, to the set of example
sentences of each synset in \UWN\ that contains at least one lemma
from \LEM.
Moreover, for each lemma in \LEM\ retrieved from a synset of \UWN\ and
found to be orthographically incorrect according to the \WB, the
system replaces it with the corrected lemma.

\begin{algorithm}[h]
\SetKwData{UW}{\UWN}
\SetKwData{W}{W}
\SetKwData{Synset}{synset}

\SetKwData{Lemmas}{\LEM}
\SetKwData{Klemmas}{\KLEM}
\SetKwData{Score}{\SCORE}
\SetKwData{Sentence}{\ASEN}
\SetKwData{UsedLemmas}{used\_lemmas}

\SetKwFunction{FlemmasSynset}{lemmas}
\SetKwFunction{Sexamples}{examples}
\SetKwFunction{NewExample}{addExample}

\SetKwFunction{UpdateUzWordNet}{update\_Uzwordnet}

\SetKwFor{For}{for}{do}{end}
\SetKwIF{If}{ElseIf}{Else}{if}{then}{else if}{else}{}
\SetKwFor{While}{while}{do}{}

\SetKwProg{Fn}{Function}{}{}

\Fn{\UpdateUzWordNet{\W,\Lemmas,\Sentence}}{
   
  \W $\leftarrow$ \UW\;

  \Lemmas $\leftarrow$ lemmas in \W\; $l$ lemma in \Lemmas\;
    
  \Sentence $\leftarrow$ Uzbek sentence\;
 
  \Synset $\leftarrow$ a synset in \W\;

  \FlemmasSynset(\Synset) $\leftarrow$ lemmas in \Synset\;

  \Sexamples($l$,\Synset) $\leftarrow$ examples for $l$ in \Synset\;

  \UsedLemmas$\leftarrow \emptyset$\;

  \For{\bf each $\Synset \in \W$}{
    \For{\bf each $l \in \Lemmas$}{
      \If {$l \in \FlemmasSynset(\Synset)$}{
        \NewExample(\Sentence) to \Sexamples($l$,\Synset)\;
        \UsedLemmas $\leftarrow$ \UsedLemmas $\cup\, \{ l \}$
      }
    }
\W $\leftarrow$ \W $\cup$ \Sexamples(\UsedLemmas,\Synset)\ \;

}
}
\KwRet{\W}\;

\caption{Update \UWN\ with sample sentences (Game 2).}
\label{alg-SFG-updateUW}
\end{algorithm}\noindent

\section{Related Work}
\label{s-RW}

To the best of our knowledge, none of the existing educational platforms for
natural language learning, whether or not they use 
generative AI or game elements, include Uzbek as an option; see
\cite{Cassie2023duolingo,Essafi2024jct,Karasimos2022rpll}
(multilingual); \cite{Sari2022jrt} (English, Korean);
\cite{Sofa2022taq} (Arabic); \cite{Eryigit2023call} (Turkish).  
Moreover, we are not aware of much related scholarly work
available \textit{in English}. Language learning,
however, like education in general, benefits from the application of
generative AI; see surveys \cite{Mittal2024ieee,Qian2025TechTrends}. 

Some of our work, specifically the design of Game 0, roots into the
\textit{AI Help Writing} feature of \KIT\ \cite{Agostini2025STIARedit},
and builds on research that correlates AI-generated teaching material
and CEFR rules of 
language comprehension \cite{Young2023ijacs}. In \cite{Young2023ijacs}
the authors discuss 
the advantages of \CHATGPT\ in teaching
English to learners who are classified in CEFR levels from ``A1'' to
``B1''. The contents generated by \CHATGPT\ are, according to
\cite{Young2023ijacs}, more suitable for these levels than higher
levels of proficiency. This motivated our choice of 
design for Game 0. 

Although there is substantial work on the use of ``gamification'' and
game mechanics to collect linguistic data or to improve lexical resources
in general, see for instance
\cite{Chamberlain2013chapter,Genovese2024onlinejcmt,Madge2024LREC-COLINGworkshop} 
and the references cited therein, we are aware of only one 
study \trl\ namely \cite{Benjamin2016GWC}, and none more recent \trl\ that
applies a game-based approach to the improvement of a wordnet of the
kind and structure considered in this paper. 
In \cite{Benjamin2016GWC}, ``players are asked to answer
targeted questions about their language, for which they receive
various rewards when their answers adhere to the consensus''. (p. 30)
The lexical data improved by tools developed within the Kamusi Project
described in \cite{Benjamin2016GWC} concern English definitions in
\PWN, and definitions in other wordnets linked to
\PWN\ \cite{Bond13ACL}. In contrast, our methods aim to improve
the lemmas and example sentences of \UWN.

\section{Conclusion, Limitations, and Future Work}
\label{s-Concl+FW}

This paper presented an educational system architecture that enables 
learners to practice the Uzbek language through game-playing.
The proposed architecture integrates \UWN\ with
generative AI 
to support both natural language learning and the enhancement of the
lexical resource in use. 
To achieve this, we followed a game-based perspective and designed
four new educational games for language learning. Two of them (Game 1
and Game 3)
employed a word base built on the largest and most authoritative
orthographic dictionary for Uzbek currently available; one game (Game
0) used a set of grammar rules to guide language learning at the
elementary and intermediate levels.

The design of three games (Game 1, Game 2, and Game 3) provided a
methodology for improving
\UWN\ and of any other lexical database structurally compatible with
\PWN, making the resulting enhancements a direct by-product of
learning through game-playing. These games are both educational and
lexical. One game (Game 0) is purely educational.

Our approach is replicable at the technical level. By using the
proposed architecture and games with different components \trl\ for
example, another LLM, a wordnet for another language, a different
orthographic dictionary, or alternative grammar rules \trl\ learners
can practice \emph{their} target language by game-playing, while
improving at the same time the wordnet used.

\subsection{Study limitations}

First, as mentioned in Remark \ref{rmk-vocabulary}, 
we could not find resources for mapping Uzbek
words to the A1, A2, and B1 language proficiency levels defined by
CEFR \cite[sec. 3.3]{CoE2020cefr}. Consequently, we assumed that
the CEFR levels assigned to English words apply to their Uzbek
translation. This assumption remains to be verified.

Second, and most importantly, although we designed four games, 
they have not yet been extensively evaluated,
neither in terms of their educational effects nor with respect to the
proposed lexical objectives.
The reasons are twofold: (1) the technical focus of this paper is on
system design from a software engineering perspective, and (2)
a critical mass of data, here also to interpret as the social impact
of engaging a sufficient number of learners to play the games and
provide feedback, has not yet been reached. 

Third, the metric used to assess how natural or correct a set of
lemmas from \UWN\ is when a learner composes an Uzbek sentence using
the lemmas (e.g., the computation of the Sentence Score in the \SFG) is
LLM-generated, so only partially accessible, transparent, and
explainable. For example, how the language model determines whether
 ``the lemmas are highly related and easily combined'' (cf. \PROMPT\ in the \SFG)? 
How the model assigns a numerical score \SCORE\ on a scale of 1 to 10
to measure the distance between a lemma written by the learner and a
lemma in the \WB?
Unfortunately, we cannot provide definitive answers to these and
similar questions related to how the language model in use computes.
Our structured prompting methodology, however, allows us to extract an 
interpretable and consistent metric that serves as a basis for
improving \UWN\ and for reliably scoring the learner's performance.   

\subsection{Future work}
\label{ss-FW}

This paper lays the groundwork for future research, building on the
limitations described to evaluate the impact and 
effectiveness of the proposed architecture in achieving the
educational and lexical objectives defined by the game design.

In particular, it would be interesting to evaluate the quality of the
responses generated by the \AIH\ across the games, or to experiment
with LLMs other than \GPT-3.5, especially open-source ones. This would
be valuable for replicating our game-based approach on other
low-resource languages and in contexts with limited economic resources.

The extension of the linguistic resources for (Northern) Uzbek language,
which we stored in three components of the architecture presented in
this paper, namely, \UWN, \WB\ and \textsc{Rules}, is
crucial to the quality of the responses the \AIH\ provides to the
learner, whichever LLM module is employed in the system's
architecture. 
On the grammatical side, one research direction to explore 
is the extension of \textsc{Rules} to include grammar rules defined in
the \textit{Mother Tongue} school textbook series
\cite{Rano2020book4,Nizomiddin2020book5,Nizomiddin2017book6}. 
These textbooks provide detailed explanations of essential 
lexical and grammatical rules of the language, and their integration
into the architecture would improve the prompt used in Game 0. 
On the lexical side, a direction of research is to replicate our
approach using the Open Multilingual Wordnet
\cite{Bond12GWC,Bond13ACL} or lexicographic datasets such as Wikionary.
Another direction is the design of multilingual games, 
in partnership within the \textit{DataScientia}
initiative\footnote{See http://datascientia.disi.unitn.it/. Accessed: 2026-02-21.}, 
leveraging the Universal Knowledge Core \cite{Giunchiglia2018CICLING}, a multilingual,
high-quality, large-scale, and diversity-aware machine-readable
lexical resource.

\bibliography{%
/Users/aa/biblio/bib/Linguistics,
/Users/aa/biblio/bib/CSEducation, 
./AKM2026,
/Users/aa/biblio/bib/myPublications}
\end{document}